\documentclass[11pt,letterpaper]{article}
\usepackage{cogsys}
\usepackage[T1]{fontenc}
\usepackage{times}
\usepackage[pdftex]{graphicx} % use this when importing PDF files

\usepackage{balance}
\usepackage{multirow}
\usepackage{amssymb}
\usepackage{pifont}
\newcommand{\cmark}{\ding{51}}%
\newcommand{\xmark}{\ding{55}}%
\usepackage{times}
\usepackage{soul}
\usepackage[table]{xcolor}
\usepackage{multicol}
\definecolor{linkcolor}{HTML}{A93C93}
\PassOptionsToPackage{hyphens}{url}
\PassOptionsToPackage{obeyspaces}{url}%
\usepackage[hidelinks, colorlinks=true, allcolors=linkcolor]{hyperref}
\usepackage{url}

\usepackage[utf8]{inputenc}
\usepackage[small]{caption}
\usepackage{graphicx}
\usepackage{amsmath}
\usepackage{amsthm}
\usepackage{booktabs}
\usepackage{algorithm}
\usepackage{algorithmic}
\usepackage[
textwidth=1.8cm,textsize=scriptsize]{todonotes}
\newcommand{\citet}[1]{\citeauthor{#1} (\citeyear{#1})}

\usepackage{listings}
\usepackage{relsize}
\newcommand{\code}{\lstinline[style=MyInline]}

\definecolor{PrologPredicate}{RGB}{0,0,200}
\definecolor{PrologOther}    {rgb}{0.1,0.1,0.1}
\definecolor{PrologVar}      {RGB}{145,032,039}
\definecolor{PrologComment}  {RGB}{169,082,044}
\definecolor{PrologString}   {RGB}{070,120,200}

\lstdefinestyle{MyInline}
{
  keywords = {},
  breaklines = true,
  breakatwhitespace=true,
 alsoletter={',.},
   basewidth = 0.42em,
  upquote = true,
  basicstyle = \relsize{-0.5}\rmfamily\color{PrologPredicate},
   moredelim = {*[s][\color{PrologPredicate}]{(}{)}},
   moredelim = {*[s][\color{PrologPredicate}]{:-}{.}},
  literate =
  {:-}{{{:\---}}}2
  {?-}{{{?\---}}}2
  {.>.}{{\#>}}4
  {.<.}{{\#<}}4
  {.=.}{{\#=}}4
  {.<>.}{{\#$\neq$}}4
  {.>=.}{{\#>=}}5
  {.=<.}{{\#=<}}5
  {=}{{=}}2
}
\lstdefinestyle{tree}
{
  moredelim = {*[s][\color{PrologString}]{:-}{.}},
  basicstyle = \relsize{-0.5}\rmfamily\color{PrologString},
  moredelim = {*[s][\color{PrologString}]{(}{)}},
  numbers=none,
  xleftmargin=0cm,
  basewidth = 0.42em,
}
\lstdefinestyle{MySCASP}
{
    xleftmargin=0.5cm,
    numberstyle=\tiny,
    numbers=left,
    stepnumber=1,
    mathescape = true,
  keywords = {},
  upquote = true,
   basicstyle = \relsize{-.5}\rmfamily\color{PrologPredicate},
   basewidth = 0.42em,
  moredelim = {*[s][\color{PrologOther}]{:-}{.}},
  moredelim = {[s][\color{PrologVar}]{(}{)}},
  commentstyle = \color{PrologComment},
  morecomment=[l]\%,
   literate =
  {.>.}{{\#>}}4
  {.<.}{{\#<}}4
  {.=.}{{\#=}}4
  {.<>.}{{\#$\neq$}}4
  {.>=.}{{\#>=}}5
  {.=<.}{{\#=<}}5
  {-}{{\---}}1
}
\usepackage{mdframed} %% Add vertical lines
\newmdenv[topline=true, rightline=false, leftline=true,
  bottomline=false,%
  linewidth=.25pt, innerleftmargin=0pt, rightmargin=-2pt,%
  innerrightmargin=2pt, skipabove=0pt, skipbelow=0pt]{mybar}

\usepackage{xstring}
\newcounter{desiderata}
\makeatletter    
\renewcommand\thedesiderata{\two@digits{\value{desiderata}}}

\makeatother
\makeatletter    
\newenvironment{numdesiderata}[3][]{%
  \medskip
  \noindent\textbf{\large\#{#1} #2: }
  \phantomsection
  \renewcommand{\@currentlabel}{\#{#1}}% Update the label text/name
  \label{des:#1}% Set the label
  \textit{#3}
}{}
\makeatother
\newenvironment{scasp}
{\par\smallskip\textbf{s(CASP):} \parindent0pt}
{\par}
\newenvironment{discussion}
{\par\smallskip\textbf{Discussion:} \parindent0pt}
{\hrulefill\par}

\usepackage{natbib}
\cogsysheading{X}{20XX}{1-6}{X/20XX}{X/20XX}
\ShortHeadings{Addressing 16+2 Desiderata with Goal-Directed Commonsense Reasoning}
              {A.\ R.\ Tudor, Y.\ Zeng, H.\ Wang, J.\ Arias and G.\ Gupta}

\begin{document} 

\title{Building Trustworthy AI by Addressing its 16+2 Desiderata\\ with Goal-Directed Commonsense Reasoning}
 
\author{Alexis R. Tudor$^1$}{alexisrenee1@gmail.com}
\author{Yankai Zeng$^1$}{yankai.zeng@utdallas.edu}
\author{Huaduo Wang$^1$}{huaduo.wang@utdallas.edu}
\author{Joaqu\'in Arias$^2$}{joaquin.arias@urjc.es}
\author{Gopal Gupta$^1$}{gupta@utdallas.edu}
\address{$^1$University of Texas at Dallas, Texas, USA}
\address{$^2$CETINIA, Universidad Rey Juan Carlos, Madrid, Spain}
\vskip 0.2in
 
\begin{abstract}
Current advances in AI and its applicability have highlighted the need to ensure its trustworthiness for legal, ethical, and even commercial reasons. Sub-symbolic machine learning algorithms, such as the LLMs, simulate reasoning but hallucinate and their decisions cannot be explained or audited (crucial aspects for trustworthiness). On the other hand, rule-based reasoners, such as Cyc, are able to provide the chain of reasoning steps but are complex and use a large number of reasoners. We propose a middle ground using s(CASP), a goal-directed constraint-based answer set programming reasoner that employs a small number of mechanisms to emulate reliable and explainable human-style commonsense reasoning. In this paper, we explain how s(CASP) supports the 16 desiderata for trustworthy AI introduced by Doug Lenat and Gary Marcus (2023), and two additional ones: inconsistency detection and the assumption of alternative worlds. To illustrate the feasibility and synergies of s(CASP), we present a range of diverse applications. 
\end{abstract}
  
%brittleness 

\section{Introduction}
\label{sec:intr-relat-work}

Although incredible advancements in artificial intelligence (AI) have been made, robustness in AI systems is still lacking. The recent AAAI study by \citet{aaai2025} found that a ``majority of respondents (76\%) assert that `scaling up current AI approaches' to yield AGI is
`unlikely' or `very unlikely' to succeed, suggesting
doubts about whether current machine
learning paradigms are sufficient for achieving
general intelligence''.
%
%that 76\% of AI professionals surveyed believed that current approaches were ``unlikely'' or ``very unlikely'' to yield human-level intelligence (often dubbed artificial general intelligence (AGI)). 
%
Despite that, the predominant approach towards AI remains pouring more resources, such as data, training time, and GPUs, into generative AI in order to produce marginally better results. We argue that what is necessary is not larger and more robust neural networks, but rather the encoding of commonsense knowledge alongside a capable reasoner. In this paradigm, deep learning becomes a tool for a reasoner to interface with, not an entity in itself. 

Researchers have agreed since the inception of AI %REFRM \cite{mccarthy1959} 
that commonsense knowledge and reasoning are important for the field. There are two fundamental components in commonsense reasoning: the knowledge base, containing all the commonsense knowledge about the problem domain, and the commonsense reasoner, which processes that knowledge to draw new conclusions.

\smallskip 
\noindent \textbf{Sub-symbolic vs.\ symbolic AI:}
The first part, the knowledge base, almost universally poses a problem for modern deep learning methods. In deep learning, knowledge is extracted from huge amounts of data. It is impossible to perform any reasonable quality assurance at the required scale. This problem is being tackled by improving the ability to train on small datasets %REFRM, see survey by \citet{rather2024}, 
and through the use of services like Amazon's Mechanical Turk to label large quantities of data with inexpensive labor. Using data that is not labeled or evaluated for quality tends to lead to difficulty with reasoning in these systems. In some domains, a strong foundational knowledge base is almost inherently machine-usable; 
%such as in physics-informed machine learning
%REFRM, see survey by \citet{hao2023}, 
%where commonsense knowledge takes the form of math equations. 
however, in others
%, such as the processing of natural language, 
the translation from knowledge to reasoning is more difficult.
%REFRM \cite{zhou2020}
Deep learning systems are incredibly robust to ingesting large quantities of data, but they do not reason and are prone to errors that would not be made by a system with sound commonsense knowledge. 
%something is solid is spoken language, avoid use in writing
%``inexpensive'' is better, ``cheap''  is colloquial
On the other hand, logic-based symbolic reasoners struggle with representing and efficiently processing large amounts of knowledge. A combination of deep learning and logic-based reasoning, a neuro-symbolic system, performs better on key metrics like explainability without sacrificing performance, see survey by \citet{yu2023}. Overall, systems that combine a robust knowledge base with a reasoner designed to utilize it, perform the best. The problem is far from solved though, as brittleness of the logic-based knowledge base and commonsense reasonng remains a challenge.

\smallskip 
\noindent \textbf{Cyc, a machine reasoner:}
One of the oldest attempts to use commonsense knowledge is the Cyc project. Cyc seeks to encode all commonsense knowledge into a knowledge base following its own proprietary ontology \citep{lenat1995}. For four decades the Cyc knowledge base has grown, gaining hand-crafted and hand-categorized axioms of knowledge covering every imaginable subject and ambiguity in the hopes of using that knowledge to cultivate true human-like intelligence. 
% Cyc contains not just facts (like that the sky is blue) but also relations that are self-explanatory for humans, such as that if a person is a citizen of the same country as another, they have the same heads of state but different heads on their bodies. Because the Cyc knowledge base is proprietary, it is difficult to say how extensive the knowledge base is, but it is safe to say that 40 years has made it vast, requiring a large amount of reasoners on top of it to extract useful knowledge. These 1,100+ reasoners were designed with Cyc in mind and are adept at navigating it to solve different problems. 

However, outside of Cycorp the reception of Cyc has been lukewarm. Cyc's philosophy has not pervaded AI research, possibly due to Cyc being a closed system, and deep learning-based AI research has moved to the forefront. Many of Cyc’s critics use it as an example to prove that symbolic AI is doomed for failure. However, as we look to the next steps for AI beyond deep learning, other symbolic approaches solve some of its greatest weaknesses and show promise for the future. We posit that purely deep learning systems require symbolic intervention to improve their trustworthiness. Explainability is fundamentally important for AI systems that are trustworthy enough for critical use%REFRM \cite{gunning2021}
.

\smallskip 
\noindent\textbf{LLMs and Trustworthiness:}
Recent work by \citet{lenat2023} discusses what is needed for truly trustworthy AI. The paper argues that Large Language Models (LLMs) are untrustworthy, unstable, and brittle, something that has continued to prove true as LLMs have become more and more ubiquitous. 
%Even modern versions of LLMs can have hallucinations that can compromise their security, as in a recent case, reported by \citet{spracklen2025}, where malicious actors hijacked hallucinated software packages to trick developers into inadvertently downloading malware. Additionally, even the technical report for GPT-4o, by \citet{openai2024}, admits that the latest in LLM technology still makes significant errors due to its statistical nature. It produces hallucinations and wrong answers a non-insignificant amount of the time, and is susceptible to ``jailbreaks''.
They posit that generative AI systems fail because they do not reason, but rather are very sophisticated pattern matching algorithms. They argue that LLMs would need to pair up with a  system like Cyc to become trustworthy. 
%However, Cyc itself still has some notable downsides that cannot be overcome simply by connecting it with an LLM.
However, Cyc is a complex and proprietary system.

\smallskip 
\noindent\textbf{Answer Set Programming:}
%Cyc is an extremely complex and proprietary system. 
For symbolic reasoning systems to succeed, simplicity must be paramount. The answer set programming (ASP) paradigm has shown how commonsense knowledge can be represented and reasoned over in an elegant way \citep{cacm-asp,gelfond-kahl,gupta22-GDE-commonsensereasoning}. 
Implementations of ASP such as the CLINGO system \citep{gebser2014clingo}, however, have a problem, namely, that they are based on grounding the program and then using a SAT solver or a procedure inspired by SAT to find answer sets. Thus, they do not scale well and are primarily used to solve combinatorial problems rather than large knowledge-based reasoning problems. 
% I remove we have built because of possible double blind review (just in case)
%To alleviate this, we have built a goal-directed ASP engine called s(CASP) that does not ground programs and that supports constructive negation. Thus, it can provide goal-directed commonsense reasoning. We provide a brief background on s(CASP) in Section \ref{sec:scasp}. 
%
To alleviate this, we propose to model commonsense reasoning with goal-directed ASP engines, which do not ground programs and support constructive negation, giving rise to what we have termed \textbf{goal-directed commonsense reasoning}. In particular, we propose the use of s(CASP), of which we provide a brief review in Section \ref{sec:scasp}.

\smallskip  
\noindent \textbf{Addressing 16+2 Desiderata:}
This paper provides a structured argument that s(CASP) plausibly satisfies the desiderata established by \citet{lenat2023} as capabilities required by general AI to be trustworthy.
%GG6Oct: commented out this part of the sentence: or can by continuing lines of current work.
%
We also discuss additional desiderata that a (goal-directed) commonsense reasoning system must satisfy: (i) being able to specify inconsistencies via global constraints, and (ii) being able to incorporate assumptions through multiple possible worlds. The s(CASP) system incorporates these additional desiderata. Given the simplicity of s(CASP), the fact that it is open-sourced, and that it can support all the desiderata, we present it as a publicly available system for building trustworthy AI in conjunction with machine learning.
These 16+2 desiderata are reviewed in Section \ref{sec:desiderata} with a brief description, related s(CASP) capabilities, and a discussion of related work. 
%
% The main contribution of this paper is to show how the 16+2 desiderata can be supported in s(CASP). 
%We show how these desiderata are crucial for supporting many applications of s(CASP).
The main contribution of this paper is to organize the capabilities of s(CASP), and present some of its  applications in terms of the desiderata for trustworthy AI. 
%As it is likely that a solution other than the continued scaling of deep learning systems will be necessary, we hope that this promotes a neuro-symbolic approach to commonsense knowledge and reasoning, and we outline our thoughts in Section \ref{sec:future}. 

\section{Background: s(CASP)}
\label{sec:scasp}

s(CASP), by \citet{arias18-iclp-scasp}, is a novel non-monotonic reasoner that evaluates Constraint Answer Set Programs without a grounding phase either before or during execution.
s(CASP) supports predicates and thus retains logical variables (and constraints) both during the execution and in the answer sets.
The operational semantics of s(CASP) rely on backward chaining, which is intuitive to follow and lends itself to generating explanations that can be translated into natural language (NL). 
The execution of an s(CASP) program is goal-directed (i.e., starts with a query), and returns partial stable models: the subsets of the stable models, defined by \citet{gelfond88}, which include only the positive or negative literals necessary to support the initial query. Any computed bindings for unbound variables in the query are also returned. The s(CASP) system supports first order predicates and terms, constructive negation, and coinduction. More details on s(CASP) are not included due to lack of space, but they can be found elsewhere \citep{arias18-iclp-scasp,gupta22-GDE-commonsensereasoning}.
% To the best of our knowledge, s(CASP) is the only system that
% exhibits the property of relevance, defined by
% \cite{pereira89:relevant-counterfactuals}.
%
s(CASP) has been used for many applications related to the representation of commonsense reasoning (including commercial ones as in \citet{mikko}). We outline the ones below that are crucial for this paper:

\noindent
\textbf{VECSR}: Virtually Embodied Common Sense Reasoning (VECSR) by
\citet{tudor2025} is a system that accepts high-level tasks and uses s(CASP) to plan and execute those tasks in a context-aware fashion in VirtualHome, a fully-embodied simulated environment.

\noindent
\textbf{Reliable Chatbots:} In the Amazon Alexa Socialbot Grand
Challenge~4, \citet{alexa} developed a conversational AI chatbot
based on two natural language understanding systems by
\citet{basu21-NLunderstanding}. Going further, \citet{zeng2024} used LLMs to translate natural language into predicates (and vice versa) and employ commonsense reasoning based on s(CASP) to build a reliable task-bot as well as a socialbot to hold a social conversation with a human.

\noindent
\textbf{Event Calculus (EC):} A reasoner on EC by
\citet{arias22-tplp-ec} has been used by \citet{hall21-hcvs-requirementEC} to model real-world avionics
systems, verify timed properties, and identify gaps in system
requirements, and by
\citet{vasicek24-earlyvalidationhighlevelrequirements} to model a medical device.

\noindent
\textbf{Justification in NL:} Has been used by
\citet{arias20-TC-explain-scasp} and \citet{chef} to bring eXplainable
Artificial Intelligence (XAI) principles to expert knowledge systems.

\noindent
\textbf{ILP systems:} Inductive Logic Programming (ILP) systems
generate logic programs from data. 
This includes the FOLD family of algorithms, the latest of which is FOLD-SE by \citet{wang2024fold} that learns stratified answer set programs.
% This includes LIME-FOLD
% proposed by \citet{farhad-iclp17} and its extensions FOLD-RM by
% \citet{wang2022}, FOLD-R++ by \citet{wang2022fold}, and FOLD-SE by
% \citet{wang2024fold}.

% ART: I commented the follwing two out to save some space because it is covered in desiderata 07 ans 02
\noindent
\textbf{s(LAW):} An administrative and judicial discretion reasoner by
\citet{arias2024}, which allows modeling of legal rules involving
ambiguity to infer conclusions and provides natural language
justifications for those conclusions.

\noindent
\textbf{Spatial reasoner:} Implemented by \citet{arias22-tplp-bim}, it
models dynamic information and restrictions in Building Information
Modeling (BIM) and paves the way to using logic-based methodologies
such as model refinement.

\section{Addressing 16+2 desiderata with s(CASP)}
\label{sec:desiderata}

\newcommand{\Ok}{{\color{green!70!black}\cmark}}
\newcommand{\No}{{\color{red!80!black}\xmark}}
\newcommand{\Partial}{{\color{orange!90!black}\bf!}}
\newcommand{\myf}[1]{{\relsize{-.5}\it #1}}
\newcommand{\mylink}[1]{{\ref{des:#1}}}
\begin{table*}[t]
\caption{Desiderata support by Cyc, Deep-Learning, and s(CASP). Including s(CASP) features with citation.}
\label{tab:evaluation}
\centering
\begin{tabular}{ll c @{\hskip 2em} c @{\hskip 2em} cr}
\toprule
  &
  & \multirow{2}{*}{Cyc}
  & \multirow{2}{*}{DL}
  & \multicolumn{2}{c}{s(CASP)}
  \\
  \cmidrule{5-6}
  &
  & 
  & 
  & \hspace{2em}~
  & \multicolumn{1}{r}{\relsize{-.5}Feature / Tool}  \\
  \midrule
\mylink{01}   & Explanation                                & \Ok     & \No       & \Ok      & \myf{Justification Trees$^{[1]}$}     \\
\mylink{02}   & Deduction                                  & \Ok     & \Partial  & \Ok      & \myf{Base s(CASP)$^{[2]}$}            \\
\mylink{03}   & Induction                                  & \Partial& \Partial  & \Ok      & \myf{Default Rules + FOLD$^{[3]}$}    \\
\mylink{04}   & Analogy                                    & \Partial& \Partial  & \Partial & \myf{Proposed LLM + s(CASP)}            \\
\mylink{05}   & Abductive Reasoning                        & \Ok     & \No       & \Ok      & \myf{Abducibles$^{[2]}$}              \\
\mylink{06}   & Theory of Mind                             & \No     & \Ok       & \Ok      & \myf{Chatbot$^{[4]}$}               \\
\mylink{07}   & Quantifier-fluency                         & \No     & \No       & \Ok      & \myf{Goal-Directed Execution$^{[2]}$} \\
\mylink{08}   & Modal-fluency                              & \Ok     & \Ok       & \Ok      & \myf{Deontic Logic$^{[5]}$}           \\
\mylink{09}   & Defeasibility                              & \Ok     & \Ok       & \Ok      & \myf{Default Rules + FOLD$^{[3]}$}    \\
\mylink{10}   & Pro and Con Arguments                      & \Ok     & \No       & \Ok      & \myf{ASP Multiple Worlds$^{[6]}$}     \\
\mylink{11}   & Contexts                                   & \Ok     & \No       & \Ok      & \myf{VECSR$^{[7]}$}                   \\
\mylink{12}   & Meta-knowledge and Meta-reasoning          & \No     & \No       & \Ok      & \myf{Justification Trees$^{[1]}$}     \\
\mylink{13}   & Explicitly Ethical                         & \Partial& \No       & \Ok      & \myf{Constraints$^{[2]}$}             \\
\mylink{14}   & Sufficient Speed                           & \No     & \Partial  & \Ok      & \myf{VECSR$^{[7]}$}                   \\
\mylink{15}   & Sufficiently Lingual and Embodied          & \No     & \Ok       & \Ok      & \myf{Chatbot$^{[4]}$}               \\
\mylink{16}   & Broadly and Deeply Knowledgeable            & \No     & \Partial  & \Ok      & \myf{FOLD, Chatbot, VECSR$^{[3,4,7]}$}  \\[.5em]
\mylink{17} & Inconsistency Detection                      & \Partial     & \No       & \Ok      & \myf{Constraints$^{[2]}$}             \\
\mylink{18} & Multiple Possible Worlds             & \Partial& \No       & \Ok      & \myf{ASP Multiple Worlds$^{[6]}$}     \\
  \bottomrule
\end{tabular}\\[.5em]
\centerline{\relsize{-.5} Note: \Ok\ means fully accomplished, \Partial\ means partially accomplished, and \No\ means non-achieved.}
\centerline{[1] \citet{arias20-TC-explain-scasp}, [2] \citet{arias18-iclp-scasp}, [3] \citet{wang2024fold}, [4] \citet{zeng2024},}
\centerline{[5] \citet{deontic}, [6] \citet{gelfond88}, [7] \citet{tudor2025}}
\end{table*}

% The 16 desiderata for trustworthy AI are described in detail by \citet{lenat2023}, and so we will only briefly summarize the desiderata here. We provide an analysis of how s(CASP) systems perform against these desiderata and a comparison to other work (primarily Cyc and deep learning systems). While a system that combines an LLM with Cyc could meet many of the desiderata, there are several desiderata that neither system performs suitably well on, and furthermore some of the advantages of each system would be reduced by partnership with the other. We discuss how s(CASP) successfully combines many of the benefits of both symbolic and neural AI systems. Table \ref{tab:evaluation} summarizes our findings for easy reference. \todo{Table better at the end as a conclusion}

The 16 desiderata for trustworthy AI are described in detail by \citet{lenat2023}, and so this section will only briefly summarize them. Table \ref{tab:evaluation} indicates how various AI systems accomplish the desiderata for reference. 
For each of the 16+2 desiderata, we provide examples of how the s(CASP) system fulfills them. In some applications used for illustration, we describe how we successfully combine  machine learning and s(CASP)-based automated commonsense reasoning with a goal of achieving human-level intelligence at least for domain-specific tasks.
%GG6Oct: revisit the sentence above
% ART6Oct: I weakened the sentence a little bit in line with the reviewer feedback.
%To accomplish some desiderata we describe how s(CASP) successfully combines many of the benefits of both symbolic and neural AI systems.
%
%Additionally, we discuss the accomplishments of state-of-the-art systems, primarily Cyc and deep learning systems. 
%In this paper, we focus on the desiderata which are less solved by state-of-the-art systems to enable a deeper focus on how s(CASP) performs on more difficult desiderata. Therefore, desiderata which are briefly covered in this section are covered more deeply in the Appendices.

%While a system that combines an LLM with Cyc could meet many of the desiderata, there are several desiderata that neither system performs suitably well on, and furthermore some of the advantages of each system would be reduced by partnership with the other.

% \desiderata

\newcommand{\gpt}[1]{{\color{orange}#1}}

\begin{numdesiderata}[01]{Explanation} %% next argument is the description (no empty line allowed)
  {Explainability is a commonly desired trait of AI, stating that any trustworthy AI must be able to explain any solution found. Its importance is reflected in the growing field of XAI%REFRM \cite{das2020}
  .}

\begin{scasp}
    Goal-directed execution means that one poses queries that are answered by an SLD-resolution-like process, similar to Prolog, and therefore, a proof is explicitly constructed. This proof constitutes not only the explanation but also the justification of a given decision (positive or negative). In s(CASP), the justification for any query is the trace of the successful derivation, or in cases where the query fails, is the trace of the negated query. Moreover, the framework developed by \citet{arias20-TC-explain-scasp} allows the generation of justifications in natural language and navigable HTML files.
    Thus, explainability is native to s(CASP) and has been exploited in applications such as the physician advisor for chronic heart failure management by \citet{chef} and by \citet{forsante2025}.
\end{scasp}

  \begin{discussion}
    One of the great weaknesses of deep learning systems is that they are unexplainable black boxes, where it is difficult to justify a solution. The closest LLMs get is that they can be asked ``why'' an answer was given, but the explanation provided may be for a different alternative solution or the justification itself may be a hallucination. 
  \end{discussion}

\end{numdesiderata}

% MV: Deduction
% \begin{numdesiderata}[02]{Deduction}{} Deduction is one of the foundational skills of symbolic logic systems, and all symbolic systems (including s(CASP)) can accomplish it. DL systems have more difficulty with this desideratum, merely mimicking how deduction would look rather than actually performing it. More information on Deduction is in Appendix~\ref{sec:02-deduction}.
% \end{numdesiderata}

\begin{numdesiderata}[02]{Deduction}
{A foundational aspect of trustworthy AI is its capacity for human-like deduction, encompassing logical reasoning (e.g., modus ponens), arithmetic operations, exhaustive search, and the identification of contradictions or redundancies. This includes understanding logical connectives, in particular recognizing different forms of negation, i.e., not being able to conclude \code{p} is different from being able to conclude \code{p} is false. 
}

\begin{scasp}
%Deduction is one of the easier logics for symbolic systems to perform. Given premises p and $p \implies q$, we deduce q. Suppose we are given the premises that Tweety is a bird (\lstinline{bird(tweety)}), and the formula $\forall X bird(X) \implies flies(X)$. From these two premises, we can deduce that flies(tweety) holds, i.e., Tweety can fly. Deductive reasoning is easily expressed using answer set programming. If we consider Rule 1, then given $q_1$, \dots, $q_m$, not $r_1$, \dots, not $r_n$, we can deduce $p$. Essentially, the antecedent can become more complex and can include negation-as-failure (NAF) literals \cite{gupta2022}. NAF poses a problem for traditional logic solvers, but not for s(CASP), which supports both default negation and classical negation. \commin{ART: Someone make sure that last sentence is correct please.}
  Deduction is the basis of symbolic systems based on logic. Given premises \code{p} and \mbox{\code|p $\implies$ q|}, we deduce \code{q}. Suppose we are given the premises that Tweety is a bird,  \code{bird(tweety)}, and the formula \code{$\forall$ X. bird(X) $\implies$ flies(X)}. From these two premises, we can deduce that \code{flies(tweety)} holds, i.e., Tweety can fly.
  Constraint Programming adds declarative arithmetic processing to logic programming. The mortgage example by \citet{holzbaur-clpqr} allows us to reason about the relation among the principal, \code{P}, the repayment rate, \code{Mp}, and the balance owing, \code{B}, i.e.,  the  query \code{?- mortgage(P,12,,0.01,B,Mp)} returns \code{P = 6.14*R + 0.38*B}.
  Finally, to deal with negation as failure (or default negation) when we lack evidence, the stable model semantics of \cite{gelfond88} is required, e.g., the rules \code{p :- not q}\footnote{\code|P :- Q| expresses the premise \code|Q $\implies$ P|.} and \code{q :-  not p}, have two stable models \code|{p}| and \code|{q}|. The stable model semantics can be realized via answer set programming (ASP). ASP also incorporates classical (or strong) negation in addition to default negation, e.g., \code{-flies(X) :- penguin(X)} means that \code{X} does not fly if it is a penguin (\code{fly(X)} is definitely false). Strong negation is used to give an explicit proof of a predicate's falsehood. 
  
These features are integrated into s(CASP) and have been leveraged for various applications discussed earlier.
%, and for spatial reasoning \citep{arias22-tplp-bim}.
\end{scasp}

\begin{discussion}
Prior research has found that these deductive (along with inductive and abductive) reasoning skills are somewhat lacking in LLMs \citep{cheng2024,abe2025}. This is likely due in part to the fact that LLMs do not reason, but rather match patterns, and so are prone to confabulations. 
Cyc, on the other hand, provides mixed results. Cyc has over 1,000 reasoners for deductions that are each optimized for different kinds of knowledge, contexts, and solution finding. This provides a wide breadth of deduction ability, however there are disadvantages associated with having such a large number of reasoning engines. Namely, processing power required for running multiple reasoners, and overhead from meta-reasoners deciding which reasoners are needed. There is a trade-off between accuracy and speed associated with symbolic or neural methods.
\end{discussion}
\end{numdesiderata}

\begin{numdesiderata}[03]{Induction}
{Effective inductive reasoning is essential for navigating complex, uncertain environments. It involves generalizing from specific observations, such as inferring species traits, and making temporal projections where the probability of a fact holding over time often follows predictable decay curves (e.g., linear, Gaussian). Such reasoning supports adaptive decision-making, despite the inherent risk of error. This type of reasoning is the foundation for explanation-based learning, which seeks to learn more with less data based on inducing general rules from specific examples%REFRM \cite{gao2024}
. }

\begin{scasp}
%{\color{blue}Induction can also be represented in s(CASP), since ASP can model analogical, explanatory, defeasible, counterfactual, and various other types of reasoning, thanks to the presence of negation as failure and possible world semantics.\comm{to be removed?} }
%
Inductive Logic Programming (ILP), see survey by \citet{ilp}, is a sub-field of machine learning that learns interpretable logic programs from labeled data and background knowledge. Unlike statistical models, ILP uses symbolic representations and logical inference to derive hypotheses that generalize observations in a human-understandable way. This makes ILP valuable in domains requiring transparency, prior knowledge integration, and formal reasoning, such as bio-informatics, legal systems, and explainable AI. 
%
%{\color{blue}In fact, an ASP rule captures (enumerative
%\footnote{Enumerative induction is an inductive method in which a conclusion is constructed based upon the number of instances that support it \cite{inductive-wikipedia}.}) inductive reasoning \cite{inductive-wikipedia} quite precisely by also stating the exceptions to an induced default rule.\comm{to be removed?}} 
%
The FOLD family of algorithms extends ILP by enabling the learning of default rules with exceptions, capturing common patterns of non-monotonic reasoning (\citet{gupta2023,wang2024fold}). Given the following data, the initially-learned rule is \code{flies(X) :- bird(X)} due to the data on ``\code{tweety}''. Then from the example ``\code{pengu}'' FOLD learns that penguins are an exception to the rule, so the rule \code{flies(X)} is refined by adding \code{not ab(X)}. 

\begin{lstlisting}
% Data
bird(tweety).     flies(tweety).     bird(pengu).      penguin(pengu).     -flies(pengu).
% Learned Rules
flies(X) :- bird(X), not ab(X).                        ab(X) :- penguin(X).
\end{lstlisting}

\noindent
 FOLD has been used to encode models learned from various real-world data into s(CASP) programs. All versions of the FOLD algorithm
 %, which we already enumerated in Section~\ref{sec:scasp}, 
 inductively learn rule-sets from datasets,
 %using default rules with exceptions
 provide a comparable accuracy to deep learning systems (particularly, the FOLD-SE system), and are completely explainable, thanks to the s(CASP)'s justification framework. The example above is a simple one for illustration purposes only, for more complete examples and formal review of using FOLD for ILP, please see \citet{gupta2023} and \citet{wang2024fold}
 %GG6Oct: discuss with Alexis
 %This form of induction can integrate both new information from outside of the system and leverage information already encoded. These methods improved efficiency, scalability, and retain hallmark interpretability.
\end{scasp}

\begin{discussion}
Regarding induction, the paper by Lenat and Marcus state that ``[\dots] by performing one step of reasoning, Cyc could generate tens of billions of new conclusions that follow from what it already knows''. This has the possibility to add many new, default-true statements to the knowledge base. This creates inductive conclusions on a large scale, leading to increased generalizability. However, learning based only on internal knowledge without bringing in new knowledge could lead to increased bias in the conclusions drawn by some of Cyc’s many reasoners due to normally imperceptible bias in the human-created axioms. 
\end{discussion}
\end{numdesiderata}

\begin{numdesiderata}[04]{Analogy}
{Analogy is the ability to map similarity between objects of often disparate meaning. This is defined as a representational mapping from a known ``source'' domain to a novel ``target'' domain \citep{hall1989}. In natural language, this commonly comes in the form of analogical phrases (``life is like a box of chocolates''). However, analogical reasoning is also used for various logic problems (such as Raven’s Progressive Matrices). Analogical reasoning 
%of both kinds have 
has been historically difficult for AI systems.
}

\begin{scasp}
Symbolic systems have been used to solve the analogical reasoning problem with various degrees of success \citep{mitchell2021}. Because these other successful systems \citep{falkenhainer1989,hofstadter1995} are often predicated on turning analogical problems into predicates that can then be reasoned over, s(CASP) could be used as a more modern reasoner with at least as much success. In addition, analogies can be automatically generated by taking a partially grounded s(CASP) rule, then ``lifting'' it by replacing constants with variables, then instantiating these variables with different constants. Thus, a concrete procedure for boiling water can be used to generate an identical procedure for boiling milk. Work is in progress to automate such analogical reasoning using s(CASP). 

In the social chatbot application by \citet{zeng2024}, analogical reasoning has been realized by  integrating s(CASP)'s reasoning with an LLM's ability to convert text into knowledge represented as logic predicates. If the LLM-based text-to-predicate translation module recognizes that the current topic is about the movie \textit{Titanic}, a s(CASP)-based analogical reasoner is triggered to find, for example, \textit{Catch Me If You Can} movie as a candidate for the next discussion topic, since \textit{Leonardo DiCaprio} leads in both.
\begin{lstlisting}
relevant_topic(A, B, 'has_same_person') :- movie(A), movie(B), person(C), act_in(C, A), act_in(C, B).
\end{lstlisting}
Additionally, LLMs can also help in generating analogical information, that is then used by the s(CASP) engine to gather analogous pieces of information:
\begin{lstlisting}
relevant_topic('Batman Dark Knight Rises', 'Titanic', 'hero sacrifices himself for others').
\end{lstlisting}

%Thus, s(CASP) can ``solve'' both analogical riddles and analogical phrases.
\end{scasp}

\begin{discussion}
LLMs have been found to perform well at responding to natural language analogical phrases \citep{musker2024}. However, in analogical reasoning, LLMs often make mistakes different than what humans would make due to a lack of deeper understanding. The ability to replicate responses to analogical queries is valuable but insufficient for true trustworthy AI. Alternatively, symbolic approaches can achieve good results on logical puzzles like RPMs, but lack the ability to integrate with natural language or make logical leaps like analogical phrases require without careful data population. In that area, Cyc can do well only on analogical phrases contained in its knowledge base. 
\end{discussion}

\end{numdesiderata}

\begin{numdesiderata}[05]{Abductive Reasoning}
{The term abduction refers to a form of reasoning that is concerned with the generation and evaluation of explanatory hypotheses. We could also think of abduction as assumptions-based reasoning. Abductive reasoning leads back from facts to a proposed explanation of those facts or assumptions that will explain that fact.  
}

\begin{scasp}
% According to Harman \cite{harman65}, abductive reasoning takes the following form:
%
% \begin{lstlisting}[style=tree]
% The fact B is observed.
% But if A were (assumed) true, B would be a matter of course.
% Hence, there is reason to suspect that A is true.
% \end{lstlisting}
%
% \noindent 
% In this form, \code{B} can be either a particular event or an empirical generalization. \code{A} serves an explanatory hypothesis and \code{B} follows from \code{A} combined with relevant background knowledge. Note that \code{A} is not necessarily true, but plausible and worthy of further validation. We can also think of \code{A} as an assumption that we must make to explain the observation \code{B}. A simple example of abductive reasoning is that one might attribute the symptoms of a common cold to a viral infection. Or, that if we assume viral infection, then no wonder the person has symptoms of a cold. 
More formally, abduction is a form of reasoning where, given the premise \code{P $\Rightarrow$ Q}, and the observation \code{Q}, one surmises (abduces) that \code{P} holds. More generally, given a theory \code{T}, an observation \code{O}, and a set of abducibles \code{A}, then \code{E} is an explanation of \code{O} (where \code{E $\subset$ A}) if:

\begin{lstlisting}[style=tree]
$T \cup E \models O$                                      $T \cup E$ is consistent 
\end{lstlisting}

We can think of abducibles \code{A} as a set of assumptions. Generally, \code{A} consists of a set of propositions such that if \code{p $\in$ A}, then there is no rule in theory \code{T} with \code{p} as its head (that is, there is no way to argue for \code{p}). We assume the theory \code{T} to be an answer set program. Under a goal-directed execution regime, an ASP system can be extended with abduction by simply adding the following rules for an abducible \code{p}:

\begin{lstlisting}
p :- not neg_p.                         neg_p :- not p. 
\end{lstlisting}

\noindent
This is automatically achieved for a predicate \code{p} that we want to declare as an abducible in the s(CASP) system through the declaration:
\code{#abducible p}.
\end{scasp}

\begin{discussion}
The size of Cyc's knowledge base works in its favor for abductive learning. Since Cyc does not necessarily need to build proofs at runtime, but relies on a huge created knowledge base, it is simple to step ``backwards'' along a reasoning tree for a function to find a plausible explanation for why something is true or not. 
%While Cyc does provide more functionality for various reasoning techniques than LLMs do, the use of a thousand different reasoners on its expansive database can be inefficient and lead to drawbacks. 
  \end{discussion}

\end{numdesiderata}

\begin{numdesiderata}[06]{Theory of Mind}
{Two of the desiderata deal purely with the way an AI system communicates with a user (this and \ref{des:15}). Theory of Mind involves building and continually updating nuanced models of conversation partners, including their knowledge, intentions, and communication styles, to interact appropriately. This helps the AI decide how terse or elaborate to be, when to ask clarifying questions, and how to adapt to ambiguity while avoiding miscommunication. Additionally, the AI must maintain a self-model to understand its capabilities, limitations, and role in the interaction.
}

\begin{scasp}
While Lenat and Marcus posed the combination of LLMs and Cyc as a potential solution for the shortcomings of both, that was purely an idea for the future. However, the neuro-symbolic approach of combining LLMs with s(CASP) has already been explored for a variety of domains \citep{zeng2023,basu21-NLunderstanding}. 
In this use case, the LLM provides a natural language semantic parser for a s(CASP) reasoner. This constrains the LLM to increase response truthfulness and fluency, and allows for easy interfacing with a symbolic system. 
Unlike other methods to improve LLM accuracy (such as chain-of-thought prompting % REFRM \cite{wei2022} 
or more intensive training), using a s(CASP) reasoner provides actual logically-grounded constraints. The s(CASP) system also determines the boundary of the ``cognition'' of AI with a dynamically maintained knowledge base, which contributes as an essential part to a self-model, i.e., to avoid hallucination and be aware that something beyond its capability exists. When the current dialogue is far beyond the task at hand it will be labeled as ``irrelevant'' by the LLM-based semantic parser, driving the conversation back to the task.
Additionally, attaching an LLM to s(CASP) allows s(CASP) to tailor its language according to the theory of mind as well as an LLM can.
\end{scasp}

\begin{discussion}
LLMs are very adaptable to the user, having been trained on numerous styles of text and speaking. Research finds that LLMs can automatically moderate their responses based on the tone (both spoken and written) of the user's input \citep{lin2024}. %yin2024
However, unconstrained LLMs by themselves can be easily ``jailbroken'' to speak in inappropriate ways \citep{chao2023}.
  \end{discussion}

\end{numdesiderata}

\begin{numdesiderata}[07]{Quantifier-fluency}
{Quantifier fluency relates to being able to reason correctly with quantifiers. Quantifiers play a crucial role in interpreting meaning of sentences. For example, as \citet{lenat2023} point out, the sentence ``Every Swede has a king'' versus the sentence ``Every Swede has a mother'' have to be interpreted appropriately.}
%{Quantifier-Fluency is the ability to understand ambiguous data, or data where information is missing. In logical systems, unknown variables are understood by being ``quantified'', that is, for every variable that is not known we assign it a value that is known. In answer set programs, this typically takes the form of grounding. However, not every variable can be grounded and not everything can be known. Truly trustworthy AI would need to be able to reason with incomplete state. }

%GG6Oct: need to fix

\begin{scasp}
%\commin{I think the example from s(LAW) works better ---there is a pattern for unknown information that matches with this desiderata}
%s(CASP) is goal-directed rather than ground, meaning it can reason in incomplete quantifiers. 
%Using abducibles in goal-directed ASP, information that is unknown can be guessed (possibly creating multiple worlds) in order to find a solution to a given problem. 
%For example, the following pattern models whether the documents we have to certify that \code{large_family} is valid or not:
\iffalse 
\begin{lstlisting}
large_family :- evidence(large_family).
neg_large_family:- evidence(-large_family).
large_family :- not evidence(large_family), not neg_large_family.
neg_large_family :- not evidence(-large_family), not large_family.
\end{lstlisting}

So we avoid introducing that information, and the reasoner would reason assuming both scenarios. Lines~1 and~2 capture the case where we have evidence for or against, and lines 3-6 create a loop so that we assume one option or the other during evaluation.
\fi 
Since s(CASP) supports predicates, quantifiers can be easily modeled. In an answer set program, all variable are universally quantified. However, negation as failure can introduce universally quantified variables in the body of the rules. Consider for example, the following definition of a bachelor.

\begin{lstlisting}
bachelor(X) :- man(X), not married_status(X). 
married_status(X) :- married(X,Y,T).  % X married Y at time T   
\end{lstlisting}

\noindent The variables \code{Y} and \code{T} are existentially quantified in the body of the second rule. The negation of \code{married_status} in the first rule leads to these variables being universally quantified in the body. That is, \code{X} is a bachelor if \code{X} is a man, and did not marry \textit{anyone} at \textit{anytime}. Thus, goals with universal quantification can appear in the body of s(CASP) programs, allowing it to incorporate both types of quantifiers. 

With respect to the example where we have to distinguish between every Swede having a king vs having a mother, such nuances can be modeled via ASP's global constraints, as illustrated below.

\begin{lstlisting}
false :- swede(X), not hasking(X). %Every Swede has a king (or, no Swede does not have a king)
hasking(X) :- king(Y, X).
false :- swede(X), swede(Y), X $\neq$ Y, king(K1, X), king(K2, Y), K1 $\neq$ K2. % All Swedes have same king
false :- king_of_sweden(K1), king_of_sweden(K2), K1 $\neq$ K2.  % There is only one King of Sweden
king(K,X) :- king_of_sweden(K), swede(X).

false :- swede(X), not hasmother(X). % Every Swede has a mother (or no Swede does not have a mother)
hasmother(X) :- mother(Y, X).
\end{lstlisting}
\end{scasp}

\begin{discussion}
The way LLMs handle quantifier fluency is by leaning on prior knowledge to make educated guesses. However, it can wrongly fill in unknown information with unrelated assumptions. Cyc also relies on the completeness of its knowledge base for reasoning, and thus cannot reason without information that is not in or deducible/inducible from its knowledge base. 
  \end{discussion}

\end{numdesiderata}

% MV: Modal-Fluency
% \begin{numdesiderata}[08]{Modal-Fluency}{} 
%     Modal-fluency, which is the ability to interpret statements involving uncertainty or belief, is challenging for AI systems because such statements involve reasoning about possibilities rather than definite truths. ASP, especially with s(CASP), provides a robust framework for modeling modal and deontic logic through possible world semantics as argued by \citet{deontic}, resolving classic paradoxes like the Chisholm paradox more effectively than current large language models (LLMs), which again only appear to use modal-logic. Modal-fluency is discussed in more detail in Appendix~\ref{sec:08-modal-fluency}.
% \end{numdesiderata}

\begin{numdesiderata}[08]{Modal-fluency}
{Modal-fluency is the ability to understand and use qualified statements.
%expressing things that may or may not be ultimately true. 
This includes statements such as ``He \textit{hopes} that it was successful'' or ``I \textit{believe} she's \textit{afraid} that it \textit{may} cost her the job'', which can be difficult for an AI system to reason over.  
}

\begin{scasp}
ASP is based on stable model semantics that allow for multiple possible worlds. The semantics of modal logics that can be used to model modal-fluency is also given in terms of possible worlds. Thus, modal logics can be elegantly modeled with ASP/s(CASP). The $\neg K$ operator in epistemic modal logic, e.g., is nothing but default negation. Note that the concepts of ``\code{p} is definitely true'', ``\code{p} may be true'', ``\code{p} is unknown'', ``\code{p} may be false'', ``\code{p} may be unknown'', ``\code{p} is definitely false'' are represented by ASP expressions``\code{p}'', ``\code{not -p}'', ``\code{not -p $\wedge$ not p}'', ``\code{not p}'', and ``\code{-p}'', respectively, where \code{-p} represents strong negation \citep{gelfond-kahl}. Deontic logic, or the modal logic of obligations, can also be naturally represented in ASP/s(CASP). An obligation (the \textbf{OB} modal operator) can be represented as ASP's global constraint. In fact, since the odd loop over negation represents a global constraint, it can be used to elegantly resolve the contrary-to-duty paradox (as well as other paradoxes) of deontic logic. Consider the following statements for a housing community: \begin{enumerate}
\item It ought to be the case that there are no dogs.

\item It ought to be the case that if there are no dogs, then there are no warning signs.

\item If there are dogs, then it ought to be the case that there are warning signs.

\item There are dogs.
\end{enumerate}

\noindent In our framework, this is simply modeled as: \hfill \scasp{dog}

\begin{lstlisting}
dog :- not -dog, not dog.
:- -dog, not -warning_sign.
:- dog, not warning_sign.
dog.
\end{lstlisting}

\noindent 
Given this program, no world will be permitted in which there is no warning sign. If we remove (4), then no worlds will be possible in which a dog is present, as now constraint (1) will kick in. This ASP encoding thus elegantly resolves the contrary-to-duty paradox (a.k.a. the Chisholm paradox). For a more detailed discussion of the way s(CASP) allows for deontic logic representation, see \citet{deontic}
 
\end{scasp}

\begin{discussion}
Although LLMs do not possess or store any knowledge separately, the amount of natural language they've internalized gives them some ability to ``understand'' complex sentences. Larger LLMs perform moderately well with modal logic, though there is still much room for improvement \citep{holliday2024}. While the LLMs can parse modal sentences, they lack a true understanding of the logic behind modal statements. As with analogical reasoning (Desideratum \ref{des:04}), replicating response patterns to modal logic does not equate to an understanding of it, which makes them prone to strange mistakes and hallucinations. 
% This can be contrasted with the fully symbolic representation of Cyc, where modal logic is represented in its hand-crafted axioms (if something is ``believed'' it means that the ``believer'' thinks it is true, for example). 
  \end{discussion}

\end{numdesiderata}

\begin{numdesiderata}[09]{Defeasibility}
{
Defeasibility represents the ability to acquire new information and change previously held beliefs, many of which were likely only true by default in the first place. This requires the ability to reconsider previous beliefs and evaluate them on their own merits against new information before deciding whether the new or old information should be kept.
}

\begin{scasp}
Defeasibility can be implemented using default rules. Default rules are directly supported in ASP and s(CASP) with the help of negation-as-failure.  
%Default rules  can be modified by induction with exceptions, further refining the criteria for deduction. 
Also, as discussed earlier, the FOLD algorithm primarily learns a model from data in the form of default rules with exceptions (as discussed in Desideratum \ref{des:03}). %By using this format, s(CASP) is able to incorporate the defeasibility desideratum. 
These learned rules not only include exception, they may also include exception to an exception,  exception an exception to an exception, and so on. This allows for a high level of fidelity when learning or representing a nuanced concept. 
%Because the form the knowledge takes is default rules in a logic program, it is easy to see what was learned and correct it if there are obvious mistakes. %Thus, when a constraint cannot be learned, it can be added to guarantee correct behavior. 

Note that default rules also allow ASP and s(CASP) to model human judgment. Consider a doctor prescribing a medicine \code{M} for disease \code{D} to patient \code{P}.
Medicine \code{M} may have a side-effect for patient \code{P}, and so the doctor has to make a judgment call before prescribing. The doctor could aggressively jump to the default conclusion and may just ignore the possibility of a side-effect (contraindication). 

\begin{lstlisting}
prescribe(M, D, P) :- cures(M, D), not contraindicated(M, P).
contraindicated(M, P) :-  has_side_effects(M, P).  
\end{lstlisting}
	
\noindent Or, the doctor could be conservative in jumping to the default:

\begin{lstlisting}
prescribe(M, D, P) :- cures(M, D), not contraindicated(M, P).
contraindicated(M, P) :-  not -has_side_effects(M, P).  
\end{lstlisting}

\noindent The rules above could be simplified and written, respectively, as:

\begin{lstlisting}
prescribe(M, D, P) :- cures(M, D), not has_side_effects(M, P).     %aggressive
prescribe(M, D, P) :- cures(M, D), -has_side_effects(M, P).        %conservative
\end{lstlisting}

The first rule states that if there is no information about a possible side-effect, then proceed with prescribing. The second rule states that the possibility of a side-effect must be ruled out before prescribing. 
The ability to represent these nuanced judgment calls using default and strong negation is a positive aspect of ASP and s(CASP). 

In the reliable chatbot mentioned in Desideratum \ref{des:06}, s(CASP) maintains the state of the conversation, tracking the progress of the task. Once new information comes, s(CASP) first checks if it is consistent with the current state and knowledge base, and then uses it to update the state. Inconsistent information will either be discarded or clarification will be sought from the human user.

\end{scasp}

\begin{discussion}
 Both LLMs and Cyc have a similar approach to defeasibility, which is to yield to human input whenever given. For Cyc, this is a part of the continual hand-grooming of its large knowledge base. When information is added that conflicts with existing information, a human de-conflicts the two either by removing one piece of information or adding additional context around it. Cyc also supports default rules. LLMs have a less permanent approach, wherein rather than removing data from its memory (a noted difficulty with LLMs \citep{blanco2025}), the LLMs generates new responses when challenged. While human input is almost always correct relative to AI output, both systems do lack the ability to critically evaluate their answers themselves \citep{antoniou2023}. 
  \end{discussion}

\end{numdesiderata}

% MV: Pro and Con Arguments
% \begin{numdesiderata}[10]{Pro and Con Arguments}{}
%     Truly explainable AI must represent the pros and cons of its chosen solution, especially in ambiguous situations. This is handled by the multiple worlds aspect of ASP mentioned in Desideratum \ref{des:18}, but more details on this specific desideratum are covered in Appendix~\ref{sec:10-pro-con}. 
% \end{numdesiderata}

\begin{numdesiderata}[10]{Pro and Con Arguments}
{Another facet of XAI represented in Lenat's desiderata is the ``pro and con arguments'' desideratum, which is a succinct description of the ability to explain why a decision may or may not be true. This covers the ambiguity of life, where things may not be objectively true but rather situationally true based on some confounding factors. Additionally, in any given situation there may be multiple answers that are correct enough.  
}

\begin{scasp}
Answer set programs inherently generate pros and cons in the form of multiple worlds. When looking at a solution in one possible world, s(CASP) outlines which factors must be true or false to support that world. Additionally, if a proof of a goal \code{g} fails, one can immediately query \code{not g} (which must succeed in s(CASP)) to understand the cause of failure \citep{murugesan24-automatingsemanticanalysisassurance}. Similarly, \textit{counterfactuals} can be computed in s(CASP) with ease \citep{sopam-padl} to examine various what-if scenarios.    

\end{scasp}

\begin{discussion}
This functionality is built into Cyc which provides arguments for and against each point it considers, along with heuristics for those arguments, at increasingly granular levels. LLMs, however, do not provide balanced pro and con arguments because they fold too easily against user disagreement (see Desideratum~\ref{des:09}). Being unable to defend a premise when challenged makes it impossible to trust a fair evaluation of its own answers \citep{wang2023}.
\end{discussion}
\end{numdesiderata}

% MV: Contexts in Appendix
% \begin{numdesiderata}[11]{Contexts}{}
%     Much like Desideratum \ref{des:10}, the management of different contexts is largely handled by the multiple world semantics of ASP. This is in contrast to DL systems  which struggle to maintain consistent context throughout use or Cyc which contains over 10,000 nested and bespoke contexts. Contexts is covered in Appendix~\ref{sec:11-context}. 
% \end{numdesiderata}

\begin{numdesiderata}[11]{Contexts}
{Contextual understanding allows an AI to interpret and adapt knowledge, behavior, and communication based on situational, cultural, and temporal factors, such as knowing when cheering is appropriate or recognizing shifting truths over time. It must reason both within a context (e.g., belief systems or cultural norms) and about contexts (e.g., deciding whether an inference holds true in multiple settings). 
}

\begin{scasp}
The multiple worlds reasoning of all ASP programs allows for the exploration of multiple contexts. 
Consider this example from Peter Norvig:
\begin{multicols}{2}
\relsize{-0.5}
\begin{enumerate}
\item People can talk.
\item Non-human animals are not able to talk.
\item Human-like cartoon characters can talk.
\item Fish can swim.
\item A fish is a non-human animal.
\item Nemo is a human-like cartoon character.
\item Nemo is a fish.
\end{enumerate}
\end{multicols}
% ART: New bit:
% Consider the problem of determining if "Nemo" can talk, which is true in a cartoon world but not in the real world.
The solution to the problem of whether Nemo can talk is difficult to represent in classical logic, but simple in ASP. 
% This problem, posed by Peter Norvig,is difficult to represent in classical logic but simple in ASP.
In ASP there would be two mutually exclusive worlds, a ``cartoon'' world where Nemo can talk, and a ``real'' world where Nemo cannot. Of course, context can also be carried as an argument of a predicate in s(CASP).
%\comm{The context avoid inconsistency: Related with \ref{des:02} and \ref{des:17}}
\end{scasp}

\begin{discussion}
Cyc contains over 10,000 contexts that are nested within other contexts to delineate topics, time periods, real or not real, and more. This allows for a robust differentiation of contexts for answers that fit the situation. However, it is still susceptible to encountering contexts not present in its handmade knowledge base and slowdowns related to having to go down multiple different context chains to provide answers. Still, this is better than LLMs which have difficulty keeping context even from a single conversation if the conversation grows too large \citep{hatalis2024}. 
\end{discussion}
\end{numdesiderata}

\begin{numdesiderata}[12]{Meta-Knowledge and Meta-Reasoning}
{Meta-knowledge and meta-reasoning enable an AI to reflect on its own knowledge, assess the reliability and source of its beliefs, and understand its strengths and limitations in performing tasks. This includes the ability to introspect, to adapt strategies mid-process, to account for changes in reasoning over time, and to critically evaluate the reliability of its sources. 
}

\begin{scasp}
Negation-as-failure gives us a form of meta reasoning because it allows us to reason about a failed proof, enabling correction. The co-inductive hypothesis set (CHS) maintained in s(CASP) \citep{arias18-iclp-scasp} keeps track of everything that has been found to be true, further allowing it to solve queries about the proof process itself. Did we solve this through assumption (co-inductive) or was the reasoning well-founded (inductive)? Meta-programming and meta-reasoning can also be explicitly incorporated in s(CASP) by making the CHS accessible to the knowledge engineer through appropriate built-ins.
\end{scasp}

\begin{discussion}
This function is available in a limited capacity in both LLMs and Cyc. LLMs can be asked whether their answer is correct or not, but are prone to over-correcting if any user feedback is given. Cyc has reasoners about its reasoning, and thus can perform meta-reasoning.
%, but because its knowledge base is human-constructed, it treats the contents of its knowledge base as facts.
\end{discussion}
\end{numdesiderata}

\begin{numdesiderata}[13]{Explicitly Ethical}
{Trustworthy AI must adhere to a transparent ethical framework, guided by strong core principles. Crucially, the AI's ethical commitments should be visible and understandable to those it affects, forming an unchanging part of its foundational contract with users.
}

\begin{scasp}
It is possible for users to add constraints to a s(CASP) program, forcing it to be explicitly ethical. This is unlike deep learning systems which are difficult to correct or add constraints that are not in the data they were trained on. \citet{vasicek24-earlyvalidationhighlevelrequirements} have used s(CASP) constraints to validate a patient controlled analgesia (PCA) pump modeled with event calculus, and \citet{arias24-sac-valueslaw} used s(CASP) in the automation of administrative processes. These two applications illustrate how ethical restrictions can be modeled as constraints. Additionally, recently \citet{deontic} demonstrated that there is 1-1 mapping between modal logics and ASP, and how norms of obligation, impermissibility, and permission of deontic modal logic can be modeled elegantly in s(CASP) through \textit{odd loop over negation}. Conditional norms as well as conflicting norms can be modeled. This expressiveness is crucial for representing ethical systems. 
Additionally, s(CASP) is fully open source, allowing for complete review of its source code. 
%This collaborative environment 
\end{scasp}

\begin{discussion}
As LLMs are still often ``jailbroken'' to provide data against the terms of their use, it is clear that this principle is failed by LLMs. As for Cyc, without access to their systems it is hard to verify if it is explicitly ethical or not. %However, therein lies a lack of transparency that does not at first glance allow that foundational ethical contract to be formed with it. 
\end{discussion}
\end{numdesiderata}

\begin{numdesiderata}[14]{Sufficient Speed}
{An effective AI system must respond to a user query in a reasonable length of time. While in rare cases a long processing time may be acceptable, in most use cases users expect answers to their queries in a matter of seconds.  
}

\begin{scasp}
Rather than spending a large time training, as with deep learning approaches, logic programming systems like s(CASP) perform their reasoning when the query is executed. Reasoning in real world environments with many objects, each having state and properties, becomes overwhelming and can result in long execution time for queries. To avoid this problem, many logic programs are kept small and domain specific. Thus these programs lack important context for reasoning.

The Virtually Embodied Common Sense Reasoning (VECSR) system is built on a foundation of s(CASP) reasoning to solve problems in simulated embodied environments \citep{tudor2025}. The advantage of connecting a s(CASP)-based  system to a virtual simulation (as with the other connections to LLMs and data mentioned above) is that there is ultimate access to the context required for decision making. However, this results in a large knowledge base and long durations for query-execution. The VECSR system employs several compile-time static analysis techniques that can bring the reasoning time in even large programs to under a second much of the time, faster than even LLMs (proven in \citet{tudor2025}). And unlike LLMs, logical constraints can be used to guarantee 
%a higher degree of 
correctness and executability.
% REFRM \cite{huang2022}
Using a unified and multi-domain reasoner like s(CASP) also eliminates the need for switching between multiple reasoners. Thus, s(CASP) can be used for real-world, real-time reasoning with a high degree of accuracy.
\end{scasp}

\begin{discussion}
Processing speed has long been a downside of  symbolic systems, like Cyc, as they must have a large knowledge base. Cyc relies on a complex series of reasoners and caching stratagems to manage speed of response. However, in this area LLMs enjoy the benefits of their upfront training time. Modern LLMs, once trained, are able to produce responses faster than a human would in most cases. However, the training of an LLM takes several months and large quantities of computing power. 
\end{discussion}
\end{numdesiderata}

% MV: Sufficiently Lingual and Embodied
% \begin{numdesiderata}[15]{Sufficiently Lingual and Embodied}{}
%     The reliable chatbot system by \citet{zeng2023} (discussed in Desiderata \ref{des:06}) enables a symbolic commonsense reasoning system, particularly s(CASP), to be as lingual as an LLM. Additionally, embodiment is achieved by the VECSR system discussed in Desiderata \ref{des:14}. Thus, a s(CASP) system is able to be sufficiently lingual and embodied for a variety of tasks. This is discussed in more detail in Appendix A.  
% \end{numdesiderata}

\begin{numdesiderata}[15]{Sufficiently Lingual and Embodied}
{Sufficiently Lingual and Embodied refers to the ability of an AI system to communicate sufficiently for the task at hand. And while in some applications, the minimum necessary linguistics may not include natural language at all (such as in systems designed to produce numerical answers for math problems), in most use-cases some natural language is optimal for human understanding. Even when answers can be provided with little language support, it is important to be able to explain ``why'' an answer was given. In most cases, knowledge and reasoning are language-independent, but a trustworthy AI will nonetheless be able to communicate well.
}

\begin{scasp}
Commonsense reasoning systems, including s(CASP), represent knowledge as predicates. To communicate with humans, these predicates must be translated to and from natural language. Machine translation of languages has been shown to be done well with pattern matching. Thus, we employ LLMs to perform this task with high accuracy.
The reliable chatbots referenced in Desideratum \ref{des:06} illustrate this very well. These reliable chatbot applications \citep{zeng2023} usually target some specific tasks, and achieve the goal by collecting as much necessary information as they can from the human users under the guidance of s(CASP)'s knowledge base. Take the restaurant recommendation chatbot (\citet{zeng2023} as an example. A customer's food preferences, budget, family-friendliness requirements, etc., must be known before a restaurant recommendation is given. This is encoded as part of s(CASP)'s knowledge. The chatbot uses an LLM to translate user dialogs to gather these requirements represented as predicates, then makes a recommendation using its encoded logic. It is also able to explain in natural language why it made a specific recommendation. By using an LLM as a semantic parser, the s(CASP)-based chatbot is able to have linguistic proficiency similar to a human. 

Additionally, s(CASP) has been used in a fully embodied way by the VECSR system described in Desideratum \ref{des:14}. Using a s(CASP)-controlled agent in an simulation environment, s(CASP) can be used to reason over a wide variety of external stimuli and perform tasks in an embodied way. For complete details, see \citet{tudor2025}. 
%Under the s(CASP) constraints, these conventions can be easily coded as commonsense to lead the dialogue to reach the goal. This understanding promotes the LLM to use the correct language for the task and takes advantage of the LLM as a natural language parser.
\end{scasp}

\begin{discussion}
As mentioned in Desideratum \ref{des:06}, while LLMs are very adaptable in their manner of speech they are also prone to being jail-broken. Additionally, because of the quantity of data LLMs have consumed, they often need to be fine-tuned to cover a specific domain. Cyc can cover a variety of domains and does have some functionality for natural language, but it is noted to be rudimentary. In our opinion, the best approach to making AI lingual is to leverage LLMs as translation devices for converting text into a logic-based formalism and vice versa. 
\end{discussion}
\end{numdesiderata}

\begin{numdesiderata}[16]{Broadly and Deeply Knowledgeable}
{An effective AI must possess or access a broad and deep foundation of world knowledge, from commonsense to specialized domains, enabling it to interact meaningfully and contextually with people. While memorizing facts is less critical in the internet age, a trustworthy AI must skillfully retrieve, understand, and reason about information in real time, drawing on structured sources and web services. Importantly, it should interpret the meaning, trustworthiness, and implications of knowledge, using logical, analogical, and inferential reasoning as humans do.
}

\begin{scasp}
There are multiple ways to incorporate new information into a s(CASP) system. The FOLD algorithms can induce from multiple forms of data, including traditional datasets and even images \citep{padalkar2024}. NeSyFOLD in particular demonstrates how the advantages of deep learning (such as robust image analysis) can be leveraged with s(CASP) to add explainability and transparent decision analysis to AI systems. The ability to learn from data is critical to build a system that is broadly and deeply knowledgeable, as there is too much information available in the connected modern age to ever encode by hand. Data can also be added from an LLM as in the reliable chatbot research by \citet{zeng2024} or from simulated environments as in VECSR and reasoned over using the s(CASP) system.
\end{scasp}

\begin{discussion}
LLMs consume vast quantities of data, however there is a growing concern that LLMs will run out of data before achieving true reasoning \citep{villalobos2022}. Because of the large quantity of data LLMs must be trained on, there can often be incorrect or low-quality data included, which is then reflected in wrong answers provided by the LLM. Cyc's solution to this has historically been the creation of its bespoke, human-groomed knowledge base, which trades man-hours for accuracy. However, it is not possible to encode the breadth of knowledge LLMs gain through training by hand in an amount of time that allows for it to be kept up to date with the changing world. 
%Some compromise must be made to ensure reasonable accuracy of the data within a reasonable amount of time. 
\end{discussion}
\end{numdesiderata}

\begin{numdesiderata}[17]{Inconsistency Detection}
{An effective AI system must be able to detect inconsistencies and enforce invariants, something that is easy for humans. The AI system, for example, should be able to tell that a person cannot sit and stand at the same time (inconsistency) or that an alive human must always breathe (invariant).
}

\begin{scasp}
Inconsistencies and invariants can be modeled quite simply as global constraints in ASP, and are supported in s(CASP). Examples above will be coded as:

\begin{lstlisting}
false :- person(X), sit(X), stand(X).
\end{lstlisting}

\noindent These constraints may be used in sophisticated reasoning tasks, such as if we ask a person sitting on a chair to turn on a light switch at the other end of the room, then we know that the person must stand up first.
\end{scasp}

\begin{discussion}
The ability to specify inconsistencies has not been presented as a desideratum by \citet{lenat2023}. However, it is an extensively used feature in commonsense reasoning. Humans learn a vast number of such inconsistencies and invariants and use them in reasoning. LLMs do not have an explicit representation of inconsistencies, and the lack of enforcing consistency is a source of hallucinations. Cyc's reasoners have to have inconsistency detection embedded, and it is likely that the various reasoners incorporate it. 
%Detection of such inconsistencies has played an important role in automating system assurance \cite{anitha-iclp}, for example. System assurance experts create a database of inconsistencies and invariants against which claims that a system performs correctly are checked. 
\end{discussion}
\end{numdesiderata}

\begin{numdesiderata}[18]{Multiple Possible Worlds}
{
An effective AI system should reason about possible worlds that can simultaneously exist. Humans have the ability to simultaneously represent multiple possible worlds in their mind and reason over each separately. For example, the real world and the world of cartoons are two different worlds with parts that are common (fish can swim in both worlds, for example) and certain parts that are inconsistent (fish talk like humans in the cartoon world and not in the real world). 
}

\begin{scasp}
In s(CASP), multiple possible worlds are easily represented through even loops over negation as discussed in the Abduction and Context desiderata. s(CASP) explicitly supports co-inductive reasoning (which relies on greatest fixpoint semantics), essential to representing multiple possible worlds. 
\end{scasp}

\begin{discussion}
LLMs do not have an explicit notion of multiple possible worlds, and this again can be a source of hallucinations. In fact, machine learning systems are often not even aware of such nuances, and conflating of multiple worlds can be a source of inaccuracy and hallucination. Cyc supports contexts which can be thought of as different worlds, however, additional logic is needed to distinguish between different contexts and to indicate whether one context is a sub-context of another one. Lack of support for multiple possible worlds is a legacy of Russell's insistence on only relying on well-founded constructs \citep{gupta22-GDE-commonsensereasoning}. ASP/s(CASP) break away from this dogma and support non-well-founded semantics, crucial to supporting commonsense reasoning. 
\end{discussion}
\end{numdesiderata}

%ASP/s(CASP) can be thought of as modeling commonsense reasoning in a formalized/structured way, where Cyc is really brute force and ad hoc. 

% ART: Paragraph from an old version, leaving it here in case anyone wants to use it.
% In the paper proposing these desiderata, Lenat posits that a combination of LLMs and Cyc could fulfill them. However, it is clear that some desiderata are not inherent to either system, and what's more, integrating the two together would compromise the some desiderata that are accomplished well by one of the systems. For example, integrating Cyc with an LLM would likely at least jeopardize the speed of its responses. We posit there needs to be a new system that combines many of the advantages of both systems for future trustworthy AI. 

\section{Conclusion}
\label{sec:future}

Table~\ref{tab:evaluation} shows that the 16 (+2) desiderata that trustworthy AI systems must support could be supported by s(CASP) either directly or through recent applications.
%While ASP/s(CASP) can be thought of as modeling commonsense reasoning in a formal/structured manner, a solution like Cyc relies on its large human-created knowledge base and proprietary reasoners. Moreover, s(CASP) has been tested in combination with other tools, such as ILP, LLM, and simulators, achieving results comparable to those obtained by state-of-the-art approaches.
As future work, we believe that in order for s(CASP) to enable truly reliable AI it is necessary to improve its execution efficiency, applying techniques such as the dynamic consistency check by \citet{arias22-padl-dcc}, and to build a large foundation of commonsense knowledge like Cyc already possesses. Brittleness is another issue in symbolic knowledge representation, i.e., KB construction is error prone, and a small change can drastically change the meaning of a KB. This is somewhat alleviated by providing explanations and program debugging/tracing tools created for s(CASP). However, what will further help are tools that can statically analyze, for example, if adding a constraint will make a KB inconsistent. 
Nevertheless, s(CASP) has already proven its usefulness in specific applications, such as the IRMA system by Forsante Corp of Finland for automating the Clozapine drug delivery guidelines. IRMA is the first logic-based system certified under the EU Medical Device Regulation (class 2b) \citep{forsante2025}. Thus s(CASP) is used commercially while still being open source and freely available for anyone looking to create trustworthy AI applications.

\begin{acknowledgements} 
\noindent
We are grateful to Vinay Chaudhri for bringing the paper by Lenat and Marcus to our attention.
\end{acknowledgements} 

%\vspace{-0.25in}

{\parindent -10pt\leftskip 10pt\noindent
\bibliographystyle{cogsysapa}
\bibliography{format,scasp,computationalthinking}

}

% Leave a blank line before the closing brace to ensure the final 
% reference has the proper indentation. 

\newpage
\appendix

\end{document}